\documentclass{article}
\usepackage{spconf}
\usepackage{amsmath,amssymb,amsthm}
\usepackage{dsfont}
\usepackage{graphicx}
\usepackage{booktabs}
\usepackage{algorithm}
\usepackage{algpseudocode}
\usepackage{cite}
\usepackage[colorlinks=false]{hyperref}

\usepackage{eso-pic}
\AddToShipoutPictureFG*{\AtPageUpperLeft{\raisebox{-0.55in}{%
  \makebox[\paperwidth]{\footnotesize This work has been submitted to
  the IEEE for possible publication. Copyright may be transferred
  without notice, after which this version may no longer be
  accessible.}}}}



\newtheorem{lemma}{Lemma}
\newtheorem{proposition}{Proposition}

\newcommand{\Ls}{\mathcal{L}_s}
\newcommand{\Real}{\mathbb{R}}

\begin{document}
\ninept

\title{Score-Based Diffusion for Hypergraph Signals in the
\lowercase{t}-Spectral Domain}

\name{Zhiyuan Dong, Carlos Mundo-Levano, Daniel L. Lau, Gonzalo R.
Arce\thanks{Z. Dong is with the Institute for Financial Services
Analytics, University of Delaware, Newark, DE 19716, USA.}\thanks{C.
Mundo-Levano and G. R. Arce are with the Department of Electrical and
Computer Engineering, University of Delaware, Newark, DE 19716,
USA.}\thanks{D. L. Lau is with the Department of Electrical and
Computer Engineering, University of Kentucky, Lexington, KY 40506,
USA.}\thanks{This work was supported in part by the U.S. National
Science Foundation under Cooperative Agreement HAIC No. 2433450, by
NSF Grants CIF 2230161, CIF 2617115, IIS 2546201, and by the
Institute of Financial Services Analytics, co-sponsored by JP Morgan
Chase \& Co.}}
\address{Institute for Financial Services Analytics, and\\
Department of Electrical and Computer Engineering,
University of Delaware\\
Department of Electrical and Computer Engineering,
University of Kentucky}

\maketitle

\begin{abstract}
We introduce the first score-based generative model for signals
\emph{on} a fixed hypergraph, formulated in the spectral domain of
t-product hypergraph signal processing (t-HGSP). The naive tensor
heat semigroup is ill-posed, since after palindromic symmetrization
the spectral slices of the Laplacian tensor are no longer positive
semidefinite. A
per-mode frequency shift repairs it, yielding a forward Ornstein--Uhlenbeck diffusion
that decouples per (eigenmode, slice) pair with closed-form marginals,
exact score targets, and an exactly isotropic stationary law. A
lightweight slice-conditioned denoiser is trained on the resulting
real spectral grid. Sampling combines its score on data-active modes
with the closed-form score elsewhere, which proves essential for
stability at scale. A closed cosine-form observation operator turns
missing-node recovery into linear-Gaussian guidance with stabilized
posterior sampling. On hypergraphs of up to 500 nodes, our model beats
a clique-expansion baseline by margins growing from $2\times$ to
$27\times$ in covariance error, matches the graph-spectral statistics
of the data better than the graph-basis model itself, and reaches the
finite-sample estimation floor on data rotated off the t-spectral
axes. Posterior sampling reduces missing-node NMSE to $0.28$--$0.64$
(classical baselines $0.87$--$1.1$) at conservative coverage, and
cuts the framework's own denoising baseline by up to $3.6\times$.
\end{abstract}


\section{Introduction}
Score-based diffusion models are the state of the art for generative
modeling and for solving inverse problems through posterior
sampling~\cite{song2021sde,ho2020ddpm,chung2023dps}. Extending them
beyond Euclidean grids is an active frontier: on graphs, generative
diffusion for \emph{signals} has been studied with white-noise
forwards~\cite{ugnn} and with heat-driven forwards that exploit the
graph spectrum~\cite{gad}. Many real interactions, however, are
intrinsically multiway, as in co-authorship, traffic corridors, and
biochemical reactions, and are only crudely captured by pairwise graphs.
Hypergraph signal processing based on the t-product tensor algebra
(t-HGSP)~\cite{penapena2023} preserves this higher-order structure: it
equips a hypergraph with a symmetric tensor shift, a spectral
decomposition, and a sampling theory directly analogous to classical
GSP.

This work fills the remaining quadrant: generative diffusion for
\emph{hypergraph signals} driven by the \emph{t-spectral} heat
semigroup. Diffusion has been applied to hypergraph \emph{structure}
generation~\cite{hygene,hedge}, and concurrent work~\cite{hedge} even
uses a heat--OU forward on relaxed incidence matrices. Our task
(signals on a fixed hypergraph) and algebra (t-product, per-slice
spectra) are orthogonal to that line.

Our contributions: \textbf{(i)} a lemma showing the
na\"ive t-heat flow is ill-posed, with a per-mode frequency shift that
repairs it while preserving the spectral ordering (Sec.~\ref{sec:wp});
\textbf{(ii)} a Heat--OU forward SDE that decouples per (eigenmode,
DFT-slice) with closed-form marginals and mode-matched noise giving an
exactly isotropic stationary law (Sec.~\ref{sec:fwd});
\textbf{(iii)} a
hybrid analytic/neural score and stabilized likelihood guidance that
make sampling and posterior sampling robust at scale, with a closed
cosine-form, matrix-free observation operator linking recovery to the
t-HGSP sampling theory (Sec.~\ref{sec:sampling}); \textbf{(iv)}
single-GPU experiments up to $N{=}500$ nodes: generation gains over a
clique-expansion baseline growing from $2\times$ to $27\times$ with
$N$ (reaching the estimation floor on off-axis data), plus
conservative-calibrated missing-node recovery and denoising
(Sec.~\ref{sec:exp}).

\section{Related work}
\label{sec:rel}
\textbf{Score-based diffusion and inverse problems.} Diffusion models
learn a score along a forward noising process and sample by reversing
it~\cite{ho2020ddpm,song2021sde,vincent2011}. Linear inverse problems
are then solved training-free by likelihood-guided reverse
sampling~\cite{chung2023dps,song2023pigdm}. We inherit this machinery
but replace the isotropic forward with one generated by the hypergraph
itself.

\textbf{Structured forward processes.} On images, heat-dissipation and
blurring forwards~\cite{rissanen2023,hoogeboom2023blur} corrupt
coarse-to-fine along the Euclidean Laplacian's spectrum. Our forward is
the same idea on a fundamentally different operator, where the naive
transfer fails: the symmetrized Laplacian \emph{loses positive
semidefiniteness} (Lemma~\ref{lem:indef}), an obstruction with no Euclidean counterpart
that forces the per-mode repair of Sec.~\ref{sec:wp}.

\textbf{Generative models on graphs and hypergraphs.} For
\emph{graph signals}, diffusion generators exist with white-noise
forwards~\cite{ugnn} and graph-spectral heat forwards~\cite{gad}. Both
inherit the pairwise Laplacian, and with it the lossy collapse of
multiway interactions. Hypergraph \emph{structure} generation exists via local
expansion~\cite{hygene} and, concurrently, a heat--OU forward on
relaxed incidence matrices~\cite{hedge}. Clique-type matrix
reductions~\cite{zhou2006} underlie the graph-signal line above.
To our knowledge no prior work generates hypergraph \emph{signals}, and
none operates in the t-spectral domain~\cite{penapena2023}, whose
per-slice decoupling is what makes every component of our method
closed-form.

\section{Background: \lowercase{t}-HGSP}
\label{sec:bg}
Let $\mathcal{H}=(\mathcal{V},\mathcal{E})$, $|\mathcal{V}|=N$, with
maximum hyperedge cardinality $M$. Following~\cite{penapena2023},
$\mathcal{H}$ is represented by an $M$th-order adjacency tensor
$\mathcal{A}\in\Real^{N\times\cdots\times N}$ ($N^{M}$ entries;
permutation-normalized weights, with smaller edges padded within the
same order), a degree tensor $\mathcal{D}$, and a Laplacian
$\mathcal{L}=\mathcal{D}-\mathcal{A}$, flattened to frontal slices
($N\times N\times N^{M-2}$) and symmetrized by the palindromic map
$\mathrm{sym}(\cdot)$ into $\Ls\in\Real^{N\times N\times K}$,
$K=2N^{M-2}{+}1$. Everything below holds for arbitrary $M$. For
notational simplicity, and in all experiments, we take $M{=}3$, so
$\mathcal{A}\in\Real^{N\times N\times N}$ and $K=2N{+}1$. The t-product of
tensors $\mathcal{A}*\mathcal{B}$~\cite{kilmer2011} is the frontal-slice
circular convolution $\,(\mathcal{A}*\mathcal{B})^{(j)}=\sum_{k}
\mathcal{A}^{(k)}\mathcal{B}^{((j-k)\bmod K)}$, equivalently ordinary
matrix multiplication of every slice after a mode-3 DFT. All our
closed forms rest on this property. Palindromic symmetry makes every
DFT (``hat'') slice $\hat{L}^{(k)}$ a \emph{real symmetric} matrix, so
$\Ls$ admits a t-eigendecomposition from $K$ independent
eigendecompositions $\hat{L}^{(k)}=V_k\Lambda_kV_k^{\!\top}$, and the
t-HGFT of a signal tensor is $c_{ik}=[V_k^{\!\top}\hat{x}^{(k)}]_i$.

Because the spectrum is real-even, slices pair as $k\leftrightarrow
K{-}k$: only $K_u=(K{+}1)/2$ slices are unique, and the palindromic
signal classes of~\cite{penapena2023} have \emph{real} coefficients, so
the natural state space is the real grid $C\in\Real^{N\times K_u}$.
Multichannel node data $X\in\Real^{N\times L}$ enter this algebra by
the (linear) tube-packing of~\cite{penapena2023}. $K$-bandlimited
signals occupy the first $k_b$ lateral indices of every slice.

\section{Well-posed \lowercase{t}-spectral heat diffusion}
\label{sec:wp}
A natural forward process is the t-heat flow
$\mathrm{d}\mathcal{X}_t=-\Ls*\mathcal{X}_t\,\mathrm{d}t$. It is,
however, ill-posed:

\begin{lemma}[Loss of positive semidefiniteness]
\label{lem:indef}
Let $L_j=D_j-A_j$ be the (PSD) frontal-slice Laplacians before
symmetrization. After $\mathrm{sym}(\cdot)$, each hat-slice of $\Ls$ is
a cosine-weighted combination
$\hat{L}^{(k)}=\sum_j w_k(j)\,L_j$ with
$w_k(j)=\cos(2\pi k j/K)$-type weights taking \emph{negative} values
for $k\neq 0$. Hence non-DC hat-slices need not be PSD and can carry
negative eigenvalues (the DC slice $k{=}0$ is PSD), and the t-heat
semigroup diverges along such modes.
\end{lemma}

\emph{Proof.} The zero leading slice of $\mathrm{sym}(\cdot)$ forces
$\sum_{k=0}^{K-1}\hat{L}^{(k)}=0$, while the DC slice
$\hat{L}^{(0)}=\sum_j L_j$ is PSD and nonzero. Hence
$\sum_{k\neq0}\hat{L}^{(k)}=-\hat{L}^{(0)}\preceq0$ is nonzero, so some
hat-slice with $k\neq0$ has a negative eigenvalue.\hfill$\square$

We therefore define per-mode \emph{t-frequencies}
$f_{iu}=\lambda_{iu}-\min_{i,u}\lambda_{iu}\ \ge 0$ (a valid t-linear
shift $\Ls-\lambda_{\min}\mathcal{I}$, with the convention
$f=|\lambda|$ supported equivalently), normalized to $[0,1]$, indexed on the unique
grid $(i,u)$.
Relative corruption speed across modes, which is the inductive bias,
is preserved, and positivity is guaranteed.

\section{Closed-form Heat--OU forward}
\label{sec:fwd}
On spectral coordinates the forward SDE
$\mathrm{d}c=-a_{iu}\,c\,\mathrm{d}t+g_{iu}\,\mathrm{d}w$ decouples per
mode with rates $a_{iu}=\theta f_{iu}+\beta$. We choose
\emph{mode-matched noise} $g_{iu}^2=2a_{iu}\sigma^2$:

\begin{proposition}
\label{prop:closed}
For every mode, $c_t\,|\,c_0\sim\mathcal{N}\!\big(e^{-a t}c_0,\;
v(t)\big)$ with $v(t):=\sigma^2(1-e^{-2a t})$, and conditional score
$-(c_t-e^{-at}c_0)/v(t)$; the stationary law is exactly
$\mathcal{N}(0,\sigma^2)$ in \emph{every} mode, i.e., isotropic in
signal space.
\end{proposition}

\emph{Proof sketch:} integrate with factor $e^{at}$; the noise
integral's variance follows from It\^o isometry,
$g^2\!\int_0^t e^{-2a(t-s)}\mathrm{d}s=\tfrac{g^2}{2a}(1-e^{-2at})$,
and $g^2{=}2a\sigma^2$ makes the bracket's coefficient $\sigma^2$ for
every mode.\hfill$\square$

\begin{figure}[t]
\centering
\includegraphics[width=\columnwidth]{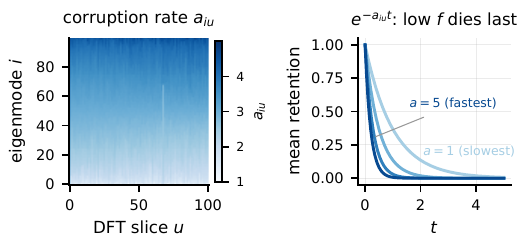}
\caption{Mode-matched forward diffusion ($N{=}100$). Left: corruption
rate $a_{iu}$ over the (eigenmode, unique-slice) grid. Right: mean
retention $e^{-a_{iu}t}$; smooth (low $t$-frequency) content survives
longest, injecting the hypergraph spectrum into the generative prior.}
\label{fig:forward}
\end{figure}

Corruption speed thus follows the hypergraph spectrum and low
t-frequencies die last (Fig.~\ref{fig:forward}), while at $T{=}5$ the
terminal law is exponentially close to the stationary Gaussian
($e^{-aT}\!\le e^{-5}$) and sampling initializes from
$\mathcal{N}(0,v(T))$. We verified
the closed forms against Euler--Maruyama simulation to floating-point
accuracy. The full pipeline is summarized in
Alg.~\ref{alg:thgdf}.

\begin{algorithm}[t]
\caption{t-spectral diffusion: training and guided sampling}
\label{alg:thgdf}
\begin{algorithmic}[1]
\State \textbf{setup:} $\Ls\!\to$ per-slice EVD $\to V,\lambda$;\;
  $f{=}\lambda{-}\lambda_{\min}$ norm.\ to $[0,1]$;\;
  $a{=}\theta f{+}\beta$;\; moments $(\mu,\varsigma^2)$, mask
  $\mathds{1}_{\text{act}}$, weights $\omega$ from training coords
\Statex \emph{--- training (DSM) ---}
\For{each step}
  \State sample $c_0\!\sim\!\text{data}$,
    $t\!\sim\!\mathcal{U}[t_{\min},T]$,
    $\epsilon\!\sim\!\mathcal{N}(0,I)$
  \State $c_t \gets e^{-at}c_0+\sqrt{v(t)}\,\epsilon$;\quad
    descend $\nabla_\phi\, \omega\|\epsilon_\phi(c_t,t)-\epsilon\|^2$
\EndFor
\Statex \emph{--- sampling (reverse SDE, log grid; uniform if guided) ---}
\State $C \gets \sqrt{v(T)}\,\mathcal{N}(0,I)$
\For{$t$ on the reverse grid}
  \State $s \gets$ hybrid score, Eq.~\eqref{eq:hybrid}
  \If{guided} \State $\hat{x}_0\!\gets$ gated Tweedie; clamp to
    $3.5\sigma$; $s \gets s+\nabla\log p(y|C)$ via
    Eq.~\eqref{eq:obs}, $\Gamma_t$; cap norm \EndIf
  \State $C \gets C + [aC + g^2 s]\,\Delta t + g\sqrt{\Delta t}\,
    \mathcal{N}(0,I)$
\EndFor
\State \Return decode $C$ via Eq.~\eqref{eq:obs} (all nodes)
\end{algorithmic}
\end{algorithm}

\subsection{Training}
The denoiser $\epsilon_\phi(C_t,t)$ is a small convolutional
residual network that treats the real $(N\times K_u)$ coefficient grid
as a single-channel image. Three inputs condition it: a sinusoidal
embedding of the diffusion time $t$ modulates every residual block
through FiLM (a learned per-channel scale and
shift)~\cite{perez2018film}, the per-position rate map $\log(1+a_{iu})$ tells each
coordinate its own signal-to-noise trajectory, and a learned
positional embedding distinguishes slices. The network totals
${\sim}0.6$M parameters and is trained by denoising score
matching~\cite{vincent2011,song2021sde},
\begin{equation*}
\min_\phi\; \mathbb{E}_{t\sim\mathcal{U}[t_{\min},T],\,c_0,\,\epsilon}
\big[\,\omega \odot \big\|\epsilon_\phi\big(e^{-at}c_0
+\sqrt{v(t)}\,\epsilon,\;t\big)-\epsilon\big\|^2\big],
\end{equation*}
so the network predicts the noise injected by the closed-form
corruption of Prop.~1, whose exact conditional score makes the
regression target free of simulation error. The minimizer recovers
the marginal score of the data distribution~\cite{vincent2011}, which
the samplers use as $s_\phi(C,t)=-\epsilon_\phi(C,t)/\sqrt{v(t)}$.
Two details matter at scale. First, the diffusion state $c_0$ is the
set of bandlimited synthesis coefficients that generate the data (the
$k_b\times K_u$ nonzero block of Sec.~\ref{sec:bg}), not the node
signals, since the node-space decode is lossy and diffusing
decoded-then-reanalyzed coordinates places the data on a dense,
non-axis-aligned subspace that destabilizes the reverse dynamics.
Second, with $k_b\ll N$ most coordinates are empty, so $\omega$
weights the active and empty groups inversely to their sizes. Each
group then carries half the loss, preventing the trivial modes from
starving the signal-bearing ones of gradient.
``Active'' means data variance above $10^{-4}$ of the maximum. Results
are insensitive to this threshold across $10^{-3}$--$10^{-5}$.

\section{Sampling and posterior sampling}
\label{sec:sampling}
\subsection{Hybrid analytic/neural score}
Reverse-time integration must keep ${\sim}97\%$ empty coordinates
contractive, since any systematic underestimate of their score
compounds as $e^{a T}$ and samples diverge (we observed $10^2$--$10^5$ blow-ups at
$N{=}500$). But on empty coordinates the marginal is Gaussian and the
score is available in closed form. We therefore use
\begin{equation}
s(C,t)=\mathds{1}_{\mathrm{act}}\!\cdot\!
s_\phi(C,t)\;+\;(1-\mathds{1}_{\mathrm{act}})\!\cdot\!
\Big[\!-\!\tfrac{C-e^{-at}\mu}{e^{-2at}\varsigma^2+v(t)}\Big],
\label{eq:hybrid}
\end{equation}
where $(\mu,\varsigma^2)$ are empirical training-coordinate moments and
$\mathds{1}_{\mathrm{act}}$ marks coordinates with non-negligible data
variance. The network is thus trusted only where the data lives. Because the
late reverse dynamics are stiff (the score scales as $1/v(t)\sim1/t$,
so explicit steps need $\Delta t\lesssim2t$), prior sampling uses a
logarithmic time grid, which improved every generation metric at equal
step count. Guided posterior sampling keeps a uniform grid, whose small
early steps the guidance loop needs.

\subsection{Missing-node recovery}
Observing channels of a node subset $S$ is \emph{linear} in the
spectral state. Using slice pairing, the entire decode chain collapses
to the closed cosine form
\begin{equation}
X_{n\ell}=\textstyle\sum_{i,u}V[n,i,u]\,w_{\ell u}\,C_{iu},\quad
w_{\ell u}=\tfrac{2 m_u}{K}\cos\!\big(\tfrac{2\pi(\ell+1)u}{K}\big),
\label{eq:obs}
\end{equation}
$m_0{=}1$, $m_{u>0}{=}2$. \emph{Derivation sketch:} the channel
read-out takes the mirror slice pair $(\ell{+}1,\,K{-}\ell{-}1)$ of the
inverse-DFT tube, whose conjugate exponentials sum to a cosine. The
real-even spectrum then makes the $k$ and $K{-}k$ terms of the DFT sum
identical, folding $K$ terms into $K_u$ with multiplicity $m_u$. The
map is therefore matrix-free, differentiable, \emph{separable} (the
hypergraph enters only through $V$; $w$ is a fixed $L\times K_u$
cosine table), and $O(1/K)$: the explicit attenuation that couples
node-space noise levels to spectral priors.

With $y=\mathrm{mask}_S(X)+n$, $n\sim\mathcal{N}(0,s_y^2I)$, we
run reverse diffusion with likelihood guidance: DPS~\cite{chung2023dps}
and $\Pi$GDM-style variance-aware guidance~\cite{song2023pigdm}
(PGDM) with
$\Gamma_t=A\,V_{0|t}A^{\!\top}+s_y^2I$ built from the empirical prior
variance.

Under the mode-matched schedule, Tweedie's
$\hat{x}_0=(C+v\,s_\phi)/e^{-at}$ amplifies neural error by up to
$e^{aT}$. We gate $\hat{x}_0$ per mode, keeping $e^{-at}\ge10^{-3}$,
which approximates exactness because the exact guidance of the
discarded modes is provably $\approx 0$. We further clamp $\hat{x}_0$ to the prior's $3.5\sigma$ box, and cap
the guidance norm at the prior-score norm. All stabilizers are
\emph{inactive} for exact scores: on conjugate Gaussian problems the
sampler reproduces the analytic posterior mean and variance (our
regression-test anchor).

\section{Experiments}
\label{sec:exp}
\textbf{Setup.} We call our model \emph{t-diff}. We use connected
random hypergraphs (a pairwise cycle
backbone plus $1.6N$ random triples) with $M{=}3$. Data are
two-cluster mixtures of $K$-bandlimited signals ($k_b=\max(4,N/25)$)
read as $L{=}3$ node channels, with 4096 training and 512 held-out
samples. The baseline is clique expansion (unit and $1/(|e|{-}1)$
weights) with the matrix-Laplacian eigenbasis, under \emph{identical}
architecture, forward machinery, budget (5k steps, EMA), and the
\emph{same} log-grid sampler. Only the spectrum differs. On held-out
node-space data we report the relative covariance error (with the
split-half data-vs-data value as the finite-sample floor), the log-PSD
correlation over data-active modes in the t-domain and the graph
domain, and the RBF-kernel maximum mean discrepancy (MMD).
\emph{Reproducibility:} the schedule is set by rule, not search
($f$ normalized to $[0,1]$, $\theta{=}4$, $\beta{=}1$, $T{=}5$ so the
slowest mode forgets to $<1\%$). Training uses AdamW, cosine LR
$2{\cdot}10^{-3}$, batch 64 (8 at $N{=}500$ for memory), and EMA 0.999
with warmup. There is no per-cell tuning, and every cell runs the same
command template.

\begin{table}[t]
\centering
\caption{Unconditional generation vs.\ held-out data (mean$\pm$std over
3 seeds; 2 seeds at $N{=}500$).}
\label{tab:e2}
\resizebox{\columnwidth}{!}{%
\begin{tabular}{clcccc}
\toprule
$N$ & model & cov.\ err.\ $\downarrow$ & t-PSD $r\uparrow$ &
g-PSD $r\uparrow$ & MMD $\downarrow$ \\
\midrule
100 & \textbf{t-diff (ours)} & 0.74$\pm$0.11 & 0.93$\pm$0.01 & 0.93$\pm$0.01 & 0.0143$\pm$0.0009 \\
 & clique ($1/(|e|{-}1)$) & 1.51$\pm$0.02 & 0.19$\pm$0.00 & -0.12$\pm$0.03 & 0.1528$\pm$0.0047 \\
 & clique (unit) & 1.47$\pm$0.02 & 0.21$\pm$0.01 & 0.00$\pm$0.03 & 0.1476$\pm$0.0036 \\
 & \emph{data floor (split-half)} & \emph{0.13} & -- & -- & -- \\
\midrule
200 & \textbf{t-diff (ours)} & 0.63$\pm$0.37 & 0.90$\pm$0.02 & 0.91$\pm$0.03 & 0.0134$\pm$0.0065 \\
 & clique ($1/(|e|{-}1)$) & 3.00$\pm$0.04 & 0.24$\pm$0.00 & 0.16$\pm$0.02 & 0.2285$\pm$0.0040 \\
 & clique (unit) & 2.86$\pm$0.04 & 0.25$\pm$0.00 & 0.18$\pm$0.02 & 0.2249$\pm$0.0042 \\
 & \emph{data floor (split-half)} & \emph{0.12} & -- & -- & -- \\
\midrule
500 & \textbf{t-diff (ours)} & 0.97$\pm$0.00 & 0.82$\pm$0.00 & 0.83$\pm$0.00 & 0.0205$\pm$0.0003 \\
 & clique ($1/(|e|{-}1)$) & 25.86$\pm$0.01 & 0.18$\pm$0.00 & -0.03$\pm$0.01 & 0.2567$\pm$0.0057 \\
 & \emph{data floor (split-half)} & \emph{0.13} & -- & -- & -- \\
 
\bottomrule
\end{tabular}}
\end{table}

\begin{figure}[t]
\centering
\includegraphics[width=\columnwidth]{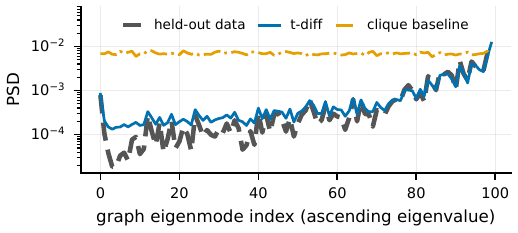}
\caption{Graph-domain power spectral density at $N{=}100$
(channel-averaged): t-diff tracks the held-out data's profile mode by
mode, while the clique baseline is flat and an order of magnitude too
energetic despite being trained in this very basis.}
\label{fig:psd}
\end{figure}

\textbf{Generation (Table~\ref{tab:e2}).} t-diff dominates every
metric at every scale. Three observations. \emph{Scaling}: the
baseline's covariance error grows with $N$ ($1.5\to3.0\to25.9$) while
t-diff stays within $5$--$8\times$ of the finite-sample floor
($0.74\to0.63\to0.97$). The t-spectral representation concentrates the
data in few coordinates, and that advantage grows with $N$.
\emph{Graph-domain win}: t-diff reproduces the \emph{graph}-spectral
profile of the data ($r=0.83$--$0.93$, Fig.~\ref{fig:psd}) far better
than the model built in that very basis ($r\approx0$). Higher-order structure is therefore
required even for graph-domain statistics. \emph{MMD}: an order of magnitude
throughout.

\textbf{Off-axis data (Table~\ref{tab:rot}).} Since the mixture is
synthesized in the t-spectral basis, we rule out basis alignment as a
confound by repeating the experiment with the $k_b$ coefficients
mixed into the span of the first $m=4k_b$ eigenmode rows by a fixed
random isometry, so the data is rank-$k_b$ but \emph{not}
sparse on t-spectral axes. t-diff is unaffected. At $N{=}100$ it
reaches the estimation floor (covariance error $0.11$ vs.\ floor
$0.12$, MMD $\approx0$), while the baseline's deficit persists. The
advantage comes from the algebra matching the interaction structure,
not from axis alignment.

\begin{table}[!t]
\centering
\caption{Non-axis-aligned variant: coefficients rotated into the
first $m$ eigenmode rows (seed 0). The unbiased MMD estimator
fluctuates around zero when the distributions match, so a slightly
negative value lies within that noise.}
\label{tab:rot}
\resizebox{\columnwidth}{!}{%
\begin{tabular}{clcccc}
\toprule
$N$ & model & cov.\ err.\ $\downarrow$ & t-PSD $r\uparrow$ &
g-PSD $r\uparrow$ & MMD $\downarrow$ \\
\midrule
100 ($m{=}16$) & \textbf{t-diff (ours)} & 0.11 & 0.95 & 0.98 & -0.0003 \\
 & clique ($1/(|e|{-}1)$) & 1.48 & 0.26 & -0.10 & 0.1745 \\
 & \emph{data floor (split-half)} & \emph{0.12} & -- & -- & -- \\
\midrule
200 ($m{=}32$) & \textbf{t-diff (ours)} & 0.88 & 0.80 & 0.81 & 0.0341 \\
 & clique ($1/(|e|{-}1)$) & 3.20 & 0.31 & 0.05 & 0.2393 \\
 & \emph{data floor (split-half)} & \emph{0.13} & -- & -- & -- \\
 
\bottomrule
\end{tabular}}
\end{table}

\begin{table}[!t]
\centering
\caption{Missing-node recovery, 50\% nodes hidden: aggregate NMSE
($\sum\|\mathrm{err}\|^2/\sum\|\mathrm{truth}\|^2$ over 8--16
trials) on hidden nodes.}
\label{tab:e3}
\begin{tabular}{lccc}
\toprule
method & $N{=}100$ & $N{=}200$ & $N{=}500$ \\
\midrule
zero-fill & 1.000 & 1.000 & 1.000 \\
low-pass~\cite{penapena2023} & 0.998 & 0.999 & 0.999 \\
Thm.~2 LS~\cite{penapena2023} & 1.064 & 0.927 & 0.992 \\
t-diff + DPS & 0.919 & 1.088 & 1.313 \\
\textbf{t-diff + PGDM} & \textbf{0.291} & \textbf{0.389} & \textbf{0.637} \\
 
\bottomrule
\end{tabular}
\end{table}

\begin{table}[!t]
\centering
\caption{Full-observation denoising: NMSE vs.\ the noisy observation
($=1.0$) at noise levels relative to signal RMS.}
\label{tab:denoise}
\begin{tabular}{llccc}
\toprule
noise & method & $N{=}100$ & $N{=}200$ & $N{=}500$ \\
\midrule
50\% & low-pass~\cite{penapena2023} & 0.37 & 0.39 & 0.37 \\
 & \textbf{t-diff + PGDM} & \textbf{0.10} & \textbf{0.18} & \textbf{0.35} \\
\midrule
100\% & low-pass~\cite{penapena2023} & 0.44 & 0.46 & 0.44 \\
 & \textbf{t-diff + PGDM} & \textbf{0.18} & \textbf{0.23} & \textbf{0.38} \\
\midrule
200\% & low-pass~\cite{penapena2023} & 0.73 & 0.76 & 0.72 \\
 & \textbf{t-diff + PGDM} & \textbf{0.31} & \textbf{0.33} & \textbf{0.47} \\
 
\bottomrule
\end{tabular}
\end{table}

\begin{figure}[!t]
\centering
\includegraphics[width=\columnwidth]{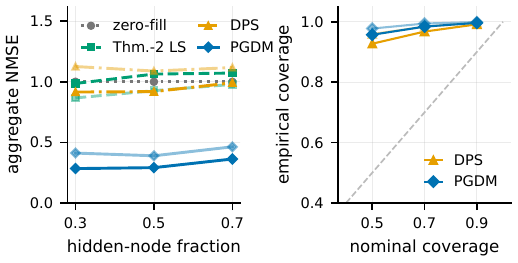}
\caption{Missing-node recovery at $N{=}100$ (saturated) and $N{=}200$
(faded). Left: NMSE vs.\ hidden fraction, where PGDM degrades
gracefully as observations vanish (low-pass overlaps zero-fill and is
omitted). Right:
empirical vs.\ nominal coverage of posterior credible intervals (both
samplers run on the t-diff prior). Intervals are conservative (above
the diagonal), never overconfident.}
\label{fig:inverse}
\end{figure}

\textbf{Inverse problems (Table~\ref{tab:e3},
Fig.~\ref{fig:inverse}).} With noise at $10\%$ of signal RMS, PGDM
guidance cuts the best classical NMSE by $3.4\times$ at $N{=}100$,
$2.4\times$ at $N{=}200$, and $1.6\times$ at $N{=}500$, and degrades
gracefully from 30\% to 70\% hidden nodes. Theorem-2
least-squares~\cite{penapena2023} is given the \emph{true} bandwidth,
yet the full spectral tubes (mode-3 fibers) the sampling theorem
operates on are not exposed by node-channel observations. This is precisely the gap a
learned prior fills, and the two are complementary rather than
competing.
\emph{Stabilizer ablation} (N{=}200, 50\% hidden): disabling the
$\alpha$-gate alone sends DPS to $3{\cdot}10^{9}$ NMSE, disabling the
norm cap alone sends PGDM to $6{\cdot}10^{2}$, and removing only
the $3.5\sigma$ clamp changes nothing ($0.380$ vs.\ $0.389$), making
the clamp defence in depth. Every guard is inactive for exact scores,
yet load-bearing for trained ones.

\textbf{Denoising (Table~\ref{tab:denoise}).} With every node observed
and noise from 50\% to 200\% of signal RMS, the natural regime of
classical spectrum shrinkage, posterior sampling still cuts the
low-pass baseline's error by $2.0$--$3.6\times$ at $N\le200$ and wins
at every scale and noise level. Together with Table~\ref{tab:e3}, the
prior wins where classical recovery works and is the only recourse
where it cannot.

Posterior credible intervals are conservative: empirical coverage is
at or above nominal at all levels (Fig.~\ref{fig:inverse}, right),
i.e., the posteriors never overclaim, at the cost of over-wide
intervals. The cause is localizable. $\Gamma_t$ uses the
\emph{mixture-wide} empirical prior variance, which books the
between-cluster spread as uncertainty even after the trajectory has
committed to a cluster. A cluster-conditional variance estimate should
tighten the intervals toward nominal and is left to future work.

\section{Conclusion}
We made t-spectral heat diffusion well-posed, exact in closed form,
stable at scale through hybrid scores and gated guidance, and useful.
It generates hypergraph signals up to $27\times$ more faithfully than
clique-expansion baselines, reaching the estimation floor off-axis,
and delivers conservative-calibrated recovery and denoising, all
reproducible on one consumer GPU.

The method is a generic prior for domains whose interactions are
multiway and whose sensors fail: traffic corridors, where missing-loop
imputation is routine (our preliminary study flags nonstationarity and
hyperedge learning as the challenges there), point clouds, where mesh
facets are literally 3-edges, and neuroimaging, where dropped EEG
channels need trustworthy error bars. Next steps are observation
models that expose the tube samples required by the sampling theorem
of~\cite{penapena2023}, cluster-conditional guidance variances, and
sequential Monte Carlo refinements of the posterior.

\bibliographystyle{IEEEbib}
\bibliography{refs}

\end{document}